\documentclass{article}
\usepackage{ijcai18}

\usepackage{times}
\usepackage{xcolor}
\usepackage{soul}
\usepackage[utf8]{inputenc}
\usepackage[small]{caption}

\usepackage{amsfonts}
\usepackage[utf8]{inputenc}
\usepackage[english]{babel}

\usepackage{amsthm}

\theoremstyle{definition}

\theoremstyle{remark}

\usepackage{latexsym}
\usepackage{amssymb}
\usepackage{enumerate}
\usepackage{amsmath}
\usepackage{tikz}
\usepackage{booktabs}
\usepackage{multirow}
\usepackage{tabularx}
\usepackage{longtable}
\usepackage{makecell}
\usepackage{graphicx}
\usepackage{subfig}

\makeatletter
\g@addto@macro\normalsize{%
 \setlength\abovedisplayskip{.3ex}
 \setlength\belowdisplayskip{.3ex}
 \setlength\abovedisplayshortskip{.3ex}
 \setlength\belowdisplayshortskip{.3ex}
}
\makeatother

\title{AAANE: Attention-based Adversarial Autoencoder for \\ Multi-scale Network Embedding}

\author{ID:3151	}

\author{
 Lei Sang\dag \ddag, Min Xu\ddag, Shengsheng Qian \S, Xindong Wu  \P \vspace*{-0.5mm}\\
 {\normalsize \dag School of Computer and Information, Hefei University of Technology} \vspace*{-1mm}\\
 {\normalsize \ddag Faculty of Engineering and IT, University of Technology Sydney} \vspace*{-1mm}\\
{\normalsize \S National Lab of Pattern Recognition Institute of Automation, Chinese Academy of Sciences} \vspace*{-1mm}\\
 {\normalsize \P  School of Computing and Informatics, University of Louisiana at Lafayette} \vspace*{-1mm}\\
 {\normalsize \texttt{lei.sang@student.uts.edu.au, Min.Xu@uts.edu.au}} \vspace*{-1mm}\\
 {\normalsize \texttt{shengsheng.qian@nlpr.ia.ac.cn, xwu@louisiana.edu}}
}

\begin{document}

\maketitle

\begin{abstract}
Network embedding represents nodes in a continuous vector space and preserves structure information from the Network. Existing methods usually adopt a  ``one-size-fits-all"  approach when concerning multi-scale structure information, such as first- and second-order proximity of nodes, ignoring the fact that different scales play  different roles  in the embedding learning. In this paper,  we propose an Attention-based Adversarial Autoencoder Network Embedding(AAANE) framework, which promotes the collaboration of different scales and lets them vote for  robust representations. The proposed AAANE consists of two components: 1) Attention-based autoencoder effectively capture the highly non-linear network structure, which can de-emphasize irrelevant scales during training. 2) An adversarial regularization guides the  autoencoder learn  robust representations by matching the posterior distribution of the latent embeddings to given prior distribution. This is the first attempt to introduce attention mechanisms to multi-scale network embedding. Experimental results on real-world networks show that  our learned attention parameters are different for every network and the proposed approach outperforms existing state-of-the-art approaches for network embedding.
\end{abstract}


\vspace*{-6mm}
\section{Introduction}

 Network embedding (NE) methods have shown outstanding performance on a number of tasks including node classification \cite{bhagat2011node,perozzi2014deepwalk}, community detection \cite{wang2017community} and link prediction \cite{lu2011link}. 
 These methods aim to learn latent, low-dimensional representations for network nodes while preserving network topology structure information.
 Networks' structures are inherently hierarchical  \cite{perozzi2017don}. As shown in Figure 1,  each individual is a member of several communities and  can be modeled by his/her neighborhoods' structure information with different scales around him/her,  
 which range from short scales structure (e.g. families, friends), to long distance scales structure (e.g. society, nation states).
 Every single scale is usually sparse and biased, and thus the node embedding learned by existing approaches may not be so robust. To obtain  comprehensive representation of a network node, multi-scale structure information should be considered collaboratively.


 \begin{figure}[]
\centering
\includegraphics[scale = 1]{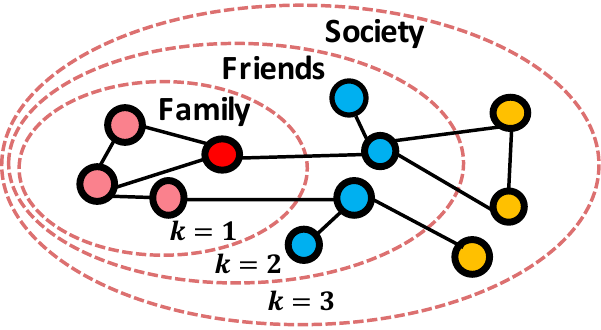}
\caption{Illustration of the multi-scale network with three scales.}
\vspace{-3mm}
\end{figure}


Recently, a number of  methods have been proposed for learning data representations from multiple scales.  
For example, DeepWalk   \cite{perozzi2014deepwalk} models multi-scale indirectly from random walk.
Line  \cite{tang2015line} proposes primarily a breadth-first strategy, sampling nodes and optimizing the likelihood independently over  short scales structure information such as 1-order and 2-order neighbors.
GraRep  \cite{cao2015grarep} generalizes LINE to incorporate information from network neighborhoods beyond 2-order, which can embed long distance scales structure information to the node representation.
More recently, some autoencoder based methods, such as DNGR \cite{cao2016deep}  and ADNE  \cite{wang2016structural}, are proposed.  DNGR  \cite{cao2016deep} learns the node embedding through stacked denoising autoencoder from the multi-scale PPMI matrix. Similarly, SDNE \cite{wang2016structural} is realized by a semi-supervised deep autoencoder model.

Despite their strong task performance, existing methods have the following limitations: 
(1) \textit{Lack of weight learning}. To learn robust and stable node embeddings, the information from multiple scales needs to be integrated. During integration, as the importance of different scales can be quite different, their weights need to be carefully decided. 
For example, if we consider very young kids  on a social network, and they may be very tightly tied to their family and  loosely tied to the  society members. 
However, for university students, they may have relatively more ties to their friends and the society than very young kids. 
Existing approaches usually assign equal weights to all scales. 
In other words, different scales are equally treated, which is not reasonable for most multi-scale networks. 
(2) \textit{Insufficient constrain for embedding distribution}. Take the autoencoder based method for example, an autoencoder is a neural network trained to attempt to copy its input to its output, 
which has a typical pipeline like ($x \rightarrow E \rightarrow z \rightarrow D \rightarrow x'$). 
Autoencoder only requires $x$ to approach $x' = D(E(x))$, and for that purpose the decoder may simply learn to reconstruct $x$ regardless of the distribution obtained from $E$. This means that $p(z)$ can be very irregular, which sometimes makes the generation of new samples difficult or even infeasible. 

In this paper, we focus on the multi-scale network embedding problem and propose a novel Attention-based Adversarial Autoencoder Network Embedding(AAANE) method to jointly capture the weighted scale structure information and learn robust representation with adversarial  regularization. 
%
We first introduce a set of scale-specific node vectors to preserve the proximities of nodes in different scales. 
The scales-specific  node embeddings are then combined for voting the robust node representations.
Specifically, our work has two major contributions. (1) To deal with the weights learning, we propose an attention-based autoencoder to infer the weights of  scales for different  nodes,  and  then capture the highly non-linear network structure, which is inspired by the recent progress of the attention mechanism  for neural machine translation  \cite{luong2015effective,qu2017attention}.
(2)  To implement regularisation of the distribution for encoded data, we  introduce adversarial training component   \cite{goodfellow2014generative} to the attention-based autoencoder, 
which can discriminatively predict whether a sample arises from the low-dimensional representations of the network or from a sampled distribution. Adversarial regularisation reduces the amount of information that may be held in the encoding, forcing the model to learn an efficient representation of the data.
Through the attention-based weight learning together with the adversarial  regularization, the proposed AAANE model can effectively combine the virtues of multiple scale information to complement and enhance each other.

%
%
We empirically evaluate the proposed AAANE approach through network visualization and node classification on benchmark datasets. The qualitative and quantitative results prove the effectiveness of our method. To the best of our knowledge, this is the first effort  to adopt the attention-based approach in the problem of multi-scale network embedding.

\begin{figure}[]
\centering
\includegraphics[scale = 0.18]{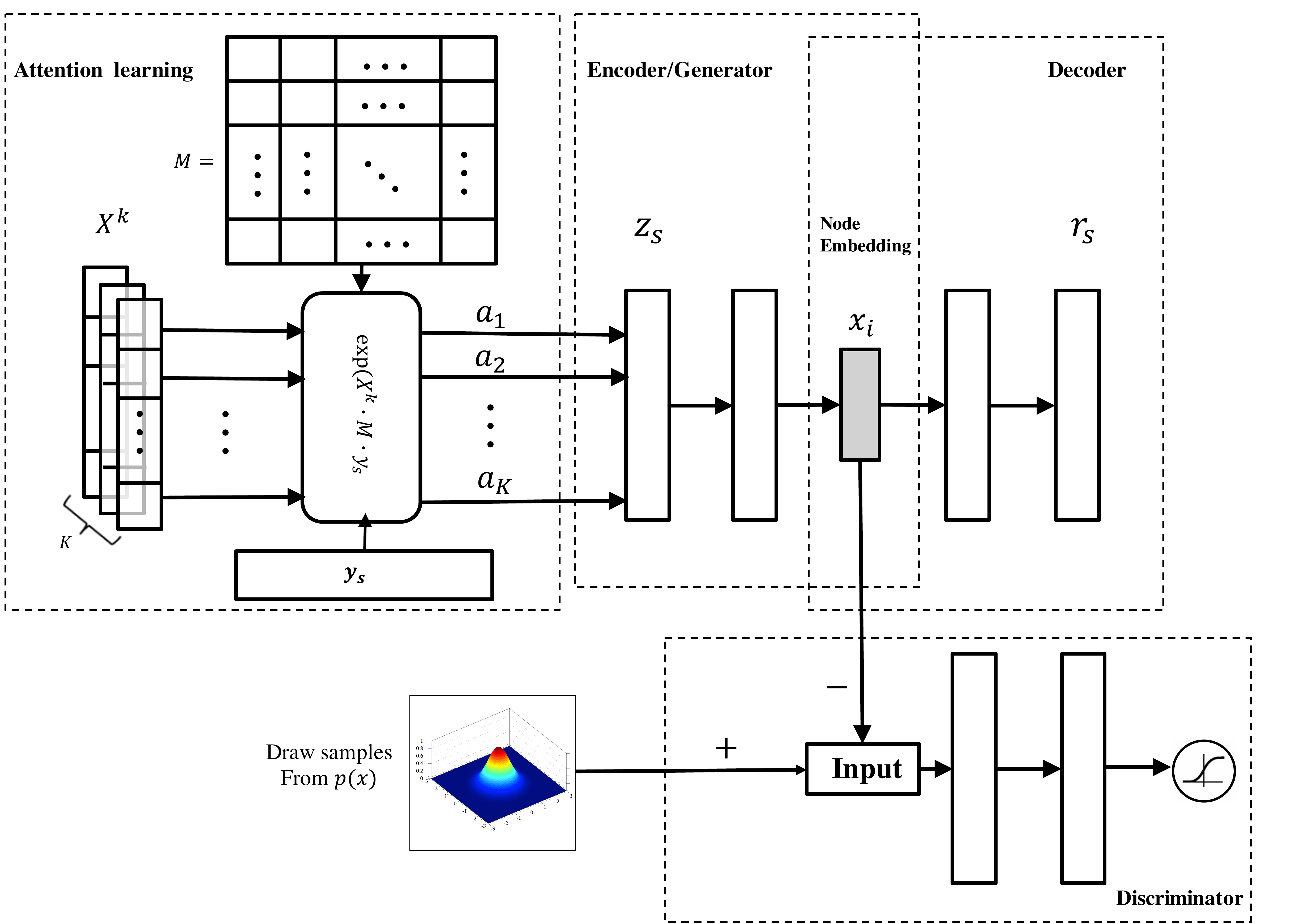}
\vspace{-6mm}
\caption{\small The architecture of  AAANE. The top row is an attention-based autoencoder that infers the weights of scales for different nodes, and then captures the highly non-linear network structure. The bottom row diagrams a discriminator  trained to discriminatively predict whether a sample arises from the hidden code of the autoencoder or from a prior distribution specified by the user. }
\vspace{-4mm}
\end{figure}
\vspace{-2mm}

\section{Preliminaries}
In this section, we briefly discuss necessary preliminaries of network embedding. 
First, we clarify the notations and formalize the problem of NE. 
Then we introduce the concept of $k$-order proximity for multi-scale embedding learning.

\subsection{Notations and Definitions}

{\textbf{Ntwork Embedding, NE}}:
An information network is represented as $G = (V, E)$ , where $V = {\{v_i\}}_{i=1,\cdots,N}$ consist a set of nodes,  $e_{i,j}=(v_i, v_j) \in E$ is an edge indicating the relationship between two nodes.   The task of NE  aims to build a low-dimensional representation $x_i \in \mathbb{R}^d $  for each node $ i \in V$, where $d$ is the dimension of embedding  space and
expected much smaller than node number  $|V|$.  
%



We  define {\em adjacency matrix} $\widetilde{A} \in \mathbb{R}^{|V| \times |V|}$for a network  and $D$ is a diagonal degree matrix.  Assuming we would like to capture the transitions from one node to another, we can define the  (first-order) probability transition matrix $A = D^{-1}\widetilde{A}$, where $A_{i, j}$  is the probability of a transition from node $v_i$ to node  $v_j$ within one step. It can be observed that the matrix $A$ is a normalized  adjacency matrix where the summation of each row equals to 1.



\vspace*{-2mm}
\subsection{Problem Definition}
In this paper, multi-scale structural information serves two functions: 1) the capture of long-distance relationship between two different vertices and    2) the consideration of distinct connections in terms of different transitional orders.

The (normalized) adjacency matrix $A$ characterizes the first-order proximity which models the local pairwise proximity between vertices.   
As discussed earlier, we believe that the $k$-order (with varying $k$) long scale relational information from the network needs to be captured when constructing such multi-scale network embedding  \cite{cao2015grarep,yang2017fast}.  To compute the various scale transition probabilities, we introduce the following $k$-order probability proximity matrix:
\vspace*{-2mm}
\begin{equation} \label{eq1}
\begin{split}
A ^k= \underbrace{ A \cdot A  \cdots A}_{k}
\end{split}
\end{equation}
where the entry  $A^k_{i, j}$  refers to  the $k$-order proximity between node $v_i$ and $v_j$.

{\textbf{Multi-Scale  Network Embedding}}: Given a network $G = (V, E)$,  the robust node representation $\{x_i\}_{v_i\in V} \subseteq R^d$ can be collaboratively learned from $k$ successively network structural information representation, $A, A^2, \ldots, A^k$, where $A^k$ captures the view of the network at scale $k$. Intuitively, each member of the family encodes a different view of social similarity, corresponding to shared membership in latent communities at differing scales.

\vspace*{-3mm}

\section{The Framework}
In this section, we first give a brief overview of the proposed AAANE, and then formulate our method of  multi-scale network embedding from  attention based adversarial autoencoder.

\vspace*{-3mm}
\subsection{An overview of the framework}
In this work, we leverage attention-based adversarial autoencoder to help learn stable and robust node embedding.   Figure 2 shows the proposed framework of Attention-based Adversarial Autoencoder for Network Embedding (\textbf{AAANE}), which mainly consists of two components, i.e., an attention-based autoencoder and an adversarial learning component.

Unlike the  matrix factory based dimension reduction method  \cite{cao2015grarep}, which maps from the original embedding space to a new space with a lower rank through a linear projection, an autoencoder can  employ non-linear activation function, such as $sigmoid$ or $tanh$, to capture complex, non-linear projections from the inputs to the outputs.  
More importantly, we introduce an attention mechanism to the autoencoder for learning the weights of structure information with different scales.
%
A standard autoencoder consists of an encoder network and a decoder network. The encoder maps the network structure information $z_s$ into a latent code $x_i$ , and the decoder reconstructs the input data as $r_s$.

Then, the adversarial learning component acts as  regularization for the autoencoder, by matching the aggregated posterior, which helps enhance the robustness of the representation $x_i$.   The generator of the adversarial network is also the encoder of the autoencoder.   The adversarial network and the autoencoder are trained jointly in two phases: the reconstruction phase and the regularization phase. In the reconstruction phase, the autoencoder updates the encoder and the decoder to minimize the reconstruction error of the inputs. In the regularization phase, the adversarial network first updates its discriminative network to tell apart the true samples (generated using the prior) from the generated samples (the hidden node embedding $x_i$ computed by the autoencoder). As a result, the proposed AAANE can jointly capture the weighted scale structure information and learn robust representations.

\vspace*{-3mm}
\subsection{Attention-based Autoencoder}

The Attention-based Autoencoder for network embedding(\textbf{AANE}) model uses a stacked neural network to preserve the structure information. As we discussed in section 2, different $k$-order proximity matrices preserve network structure information in different scales. 
Scale vector $X^k$ is column in each  $ A^k $, for $ k = 1, 2 \dots K$, which denotes the $k$-th  scale structure information for the node.  The length of each scale vector $X^k$ is the same as the node size.   The autoencoder component tries to capture the full range of structure information. We construct a vector representation $z_s$ for each node as the input of the autoencoder in the first step. 
%
In general, we expect this vector representation to capture the most relevant information with regards  to different scales of a node.
$z_s$ is defined as  the weighted summation of every scale vector $X^k, k = 1, 2, \ldots K$, corresponding to the scale index for each node.

\vspace{-5mm}
\begin{equation} \label{eq2}
\begin{split}
z_s = \sum_{k=1}^{K} a_k X^k
\end{split}
\end{equation}
For each scale vector $X^k$ of one node, we compute a positive weight $a_k$ which can be interpreted as the probability that $X_k$ is assigned by one node. Intuitively, by learning proper weights $ {a_k}$ for each node, 
%
our approach can obtain most informative scale information.
Following the recent attention based models for neural machine translation, we define the weight of scales $k$ for a node using a  softmax unit as follows:
\begin{equation} \label{eq3}
\begin{split}
a_k &= \frac{\exp(d_k)}{\sum_{j=1}^{n} \exp(d_k)} \\
d_k &= {X^k}^\top\cdot M \cdot y_s ,  \qquad  y_s = \frac{1}{K}\sum_{k=1}^{K} X^k
\end{split}
\end{equation}
where $y_s$ is  the average of different scale vector, which can capture the global context of the structure information. 
$M\in \mathbb{R}^{|V| \times |V|}$ is a matrix mapping between the global context embedding $y_s$ and each structure scale vector $X^k$, which is learned as part of the training process. 
By introducing an attentive matrix $M$, we compute the relevance of each scale vector to the node  \cite{dos2016attentive,he2017unsupervised}.   If $X^k$ and $y_s$ have a large dot product, this node believes that scale $k$ is an informative scale, i.e., the weight of scale $k$ for this node  will be largely based on the definition.

Once we obtain the weighted node vector representation $z_s \in \mathbb{R}^{|V|}$,  a stacked  autoencoder  is used to learn a low-dimensional node embedding. An autoencoder performs two actions, an encoding step, followed by a decoding step. In the encoding step, a function $f()$ is applied to the original vector representation $z_s$  in the input space and send it to a new feature space. An activation function is typically involved in this process to model the non-linearities between the two vector spaces.  At the decoding step, a reconstruction function $g()$is used to reconstruct the original input vectors  back from the latent representation space. The  $r_s$ is the reconstructed vector representation. After training, the bottleneck layer representations $x_i$  can be viewed as the low dimension embedding for the input node $v_i$.



This attention-based autoencoder is trained to minimize the reconstruction error. We adopt the contrastive max-margin objective function,  similar to
previous work   \cite{weston2011wsabie,iyyer2016feuding,he2017unsupervised}. For each input node, we randomly sample $m$ nodes from our training data as negative samples. We represent each negative sample as $n_s$,  which is computed by averaging its scale vectors as $y_s$. Our objective is to make the reconstructed embedding $r_s$ similar to the target node embedding $z_s$ while different from those negative samples $n_s$. Therefore, the unregularized objective $J$ is formulated as a hinge loss that maximizes the inner product between $r_s$ and $z_s$, and  minimizes the inner product between $r_s$ and the negative samples simultaneously:
\begin{equation} \label{eq4}
\begin{split}
J(\theta) = \sum_{s\in D}\sum_{i=1}^{m} max(0, 1-r_{s} z_{s} + r_{s}n_{i})
\end{split}
\end{equation}
where $D$ represents the training dataset.

\vspace*{-3mm}
\subsection{Adversarial Learning}

We hope to learn vector representations of the most representative scale for each  node. An autoencoder consists of two models, an encoder and a decoder, each of which has its their own set of learnable parameters. The encoder is used to get a latent code $x_i$ from the input with the constraint. The  dimension of the latent code should be less than the input dimension. The decoder  takes in this latent code and tries to reconstruct the original input. However, we argue that  training an autoencoder gives us latent codes with similar nodes being far from each other in the Euclidean space, especially when  processing noisy network data. The main reason is the insufficient constrain for embedding distribution.  Adversarial autoencoder (AAE) addresses these issues by imposing an Adversarial regularization to the bottleneck layer representation of autoencoder, and then the distribution of latent code may be shaped to match a desired prior distribution. Adversarial regularisation can reduce the amount of information that may be held in the encoding  process, forcing the model to learn an efficient representation for the network data.



AAE typically consists of a generator $G()$ and a discriminator $D()$. Our main goal is to force output of the encoder to follow a given prior distribution  $p(x)$(this can be normal, gamma .. distributions). We  use the encoder as our generator, and the discriminator to tell if the samples are from a prior distribution  or from the output of the encoder $x_i$. D and G play the following two-player minimax game with the value function V (G, D):
\begin{equation} \label{eq5}
\begin{split}
&\min_{G} \max_{D} V(D, G)  \\
&= \mathbb{E}_{p(x)} [ \log D(x_i)] + \mathbb{E}_{q(x)}[\log(1-D(x_i))]
\end{split}
\end{equation}
where $q(x)$ is the distributions of encoded data samples.


\vspace*{-2mm}
\subsection{Training Procedure}
The whole training process is done in three sequential steps: (1) The encoder and decoder are trained simultaneously to minimize the reconstruction loss of the decoder as equation \ref{eq4}. (2) The discriminator  $D$ is then trained to correctly distinguish the true input signals $x$  from the false signals $x_i$, where the $x$ is generated from target distribution, and $x_i$ is generated by the encoder by minimizing the loss function \ref{eq5}. (3) The next step will be to force the encoder to fool the discriminator by minimizing another loss function: $L = -\log(D(x_i))$. More specifically, we connect the encoder output as the input to the discriminator. Then, we fix the discriminator weights  and fix the target to 1 at the discriminator output. Later, we pass in a node to the encoder and find the discriminator output which is then used to find the loss.



\vspace*{-3mm}

\section{Experiments}
In this section, we conduct node classification on sparsely labeled networks to evaluate the performance of our proposed model.

\vspace*{-2mm}
\subsection{Datasets}
We employ the following three widely used datasets for node classification.

\textbf{Cora.}  Cora  is a research paper set constructed by  \cite{mccallum2000automating}. It contains 2, 708 machine learning papers which are categorized into 7 classes. The citation relationships among them are crawled form a popular social network.

\textbf{Citeseer.}  Citeseer is another research paper set constructed by  \cite{mccallum2000automating}. It contains 3, 312 publications and 4, 732 links among them. These papers are from 6 classes.

\textbf{Wiki.} Wiki  \cite{sen2008collective} contains 2, 405 web pages from 19 categories and 17, 981 links among them. Wiki is much denser than Cora and Citeseer.

\begin{table*}[t]
\small
\centering
\caption{Accuracy (\%) of node classification on Wiki.}
\begin{tabularx}{0.7\textwidth}{llXXXXXXXX}
\toprule
$\%$ Labeled Nodes & $10\%$ & $20\%$ &  $30\%$ & $40\%$ &$50\%$ & $60\%$ &$70\%$ & $80\%$ &$90\%$  \\
\midrule
DeepWalk          & 57.2 &62.98 &64.03 &65.78 &66.74 &68.69 &68.36 &67.85 &67.22\\
LINE                  & 57.09 &59.98 &62.47 &64.38 &66.5 &65.8 &67.31 &67.15 &65.15 \\
GraRep              & 59.55 &60.76 &62.23 &62.3 &62.76 &63.72 &63.02 &62.79 &60.17 \\
node2vec           & 58.47 &61.38 &63.9 &63.96 &66.08 &66.74 &67.73 &67.57 &66.8 \\
AIDW                 & 57.29	&61.89 &63.77	&64.26 &66.85	&67.23&69.04 &70.13&71.33 \\
\hline
AANE                & 59.95 & 64.14& 66.15& 68.40&68.66&69.34&69.25&70.89& 69.71\\
AAANE              & \textbf{60.36} & \textbf{64.98}& \textbf{67.21}&	\textbf{68.79}&\textbf{69.07}&\textbf{70.32	}& \textbf{70.85}&\textbf{72.03}& \textbf{72.45}\\
\bottomrule
\end{tabularx}
\vspace*{2mm}


\caption{Accuracy (\%) of node classification on Cora.}
\begin{tabularx}{0.7\textwidth}{llXXXXXXXX}
\toprule
$\%$ Labeled Nodes & $10\%$ & $20\%$ &  $30\%$ & $40\%$ &$50\%$ & $60\%$ &$70\%$ & $80\%$ &$90\%$  \\
\midrule
DeepWalk          & 76.37 &79.6 &80.85 &81.42 &82.35 &82.1 &82.9 &84.32 &83.39\\
LINE                   & 71.08 &76.19 &77.32 &78.4 &79.25 &79.06 &79.95 &81.92 &82.29 \\
GraRep              & 77.02 &77.95 &78.53 &79.75 &79.61 &78.78 &78.6 &78.23 &78.23 \\
node2vec           & 75.84 &78.77 &79.54 &80.86 &80.43 &80.9 &80.44 &79.7 &77.86 \\
AIDW                 & 76.21 &78.93 &80.21 &81.45.  &82.03 &82.74 &82.81 &83.69&83.92 \\
\hline
AANE  		& 77.65 & 81.50&82.49&84.43	&84.71&84.69&	84.75&85.98&86.03\\
AAANE             & \textbf{78.23} & \textbf{82.14}&\textbf{82.76}&\textbf{85.31} &\textbf{85.69}&\textbf{86.12}&	\textbf{86.02}&\textbf{86.74}&\textbf{87.21}\\
\bottomrule
\end{tabularx}
\vspace*{2mm}


\caption{Accuracy (\%) of node classification on Citeser.}
\begin{tabularx}{0.7\textwidth}{llXXXXXXXX}
\toprule
$\%$ Labeled Nodes & $10\%$ & $20\%$ &  $30\%$ & $40\%$ &$50\%$ & $60\%$ &$70\%$ & $80\%$ &$90\%$  \\
\midrule
DeepWalk       & 53.47 &54.19 &54.6 &57.55 &57 &59.02 &58.95 &58.22 &55.72\\
LINE               & 48.74 &50.87 &52.82 &52.72 &52 &52.3 &53.12 &53.54 &52.41 \\
GraRep           & 53.23 &54.34 &53.77 &54.43 &54.05 &54.57 &54.83 &55.35 &55.12 \\
node2vec        & 53.94 &54.08 &56.23 &57.34 &57.55 &60.3 &61.17 &61.24 &59.33 \\
AIDW            & 52.17&56.23&56.87	&58.26 &58.45	&59.27 &59.34	&60.38 &61.3 \\
\hline
AANE             & 55.02 & 56.15	&58.65&	58.76&58.52&	59.93&60.97&61.39&61.23\\
AAANE          & \textbf{55.45} & \textbf{56.73}	&\textbf{59.37}&\textbf{59.81}	& \textbf{60.12}	&\textbf{60.58}&	\textbf{61.43}&\textbf{61.72}& \textbf{62.38}\\
\bottomrule
\end{tabularx}
\vspace*{2mm}

\vspace*{-1mm}
\end{table*}

 \begin{figure*}[!t]
\centering
 \subfloat[DeepWalk]{\includegraphics[width=0.15\textwidth]{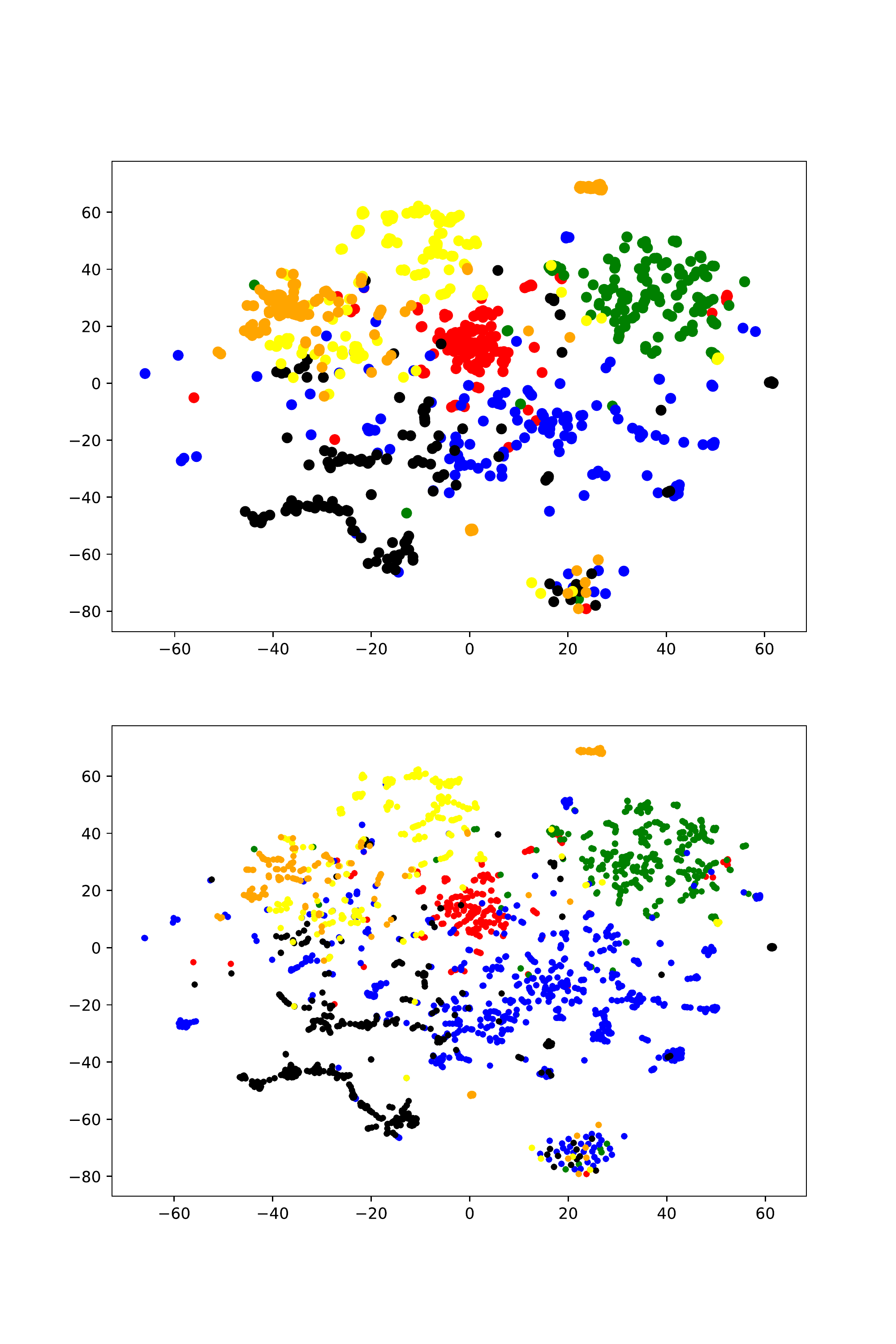}\label{fig:f1}}
\hfill
 \subfloat[LINE]{\includegraphics[width=0.15\textwidth]{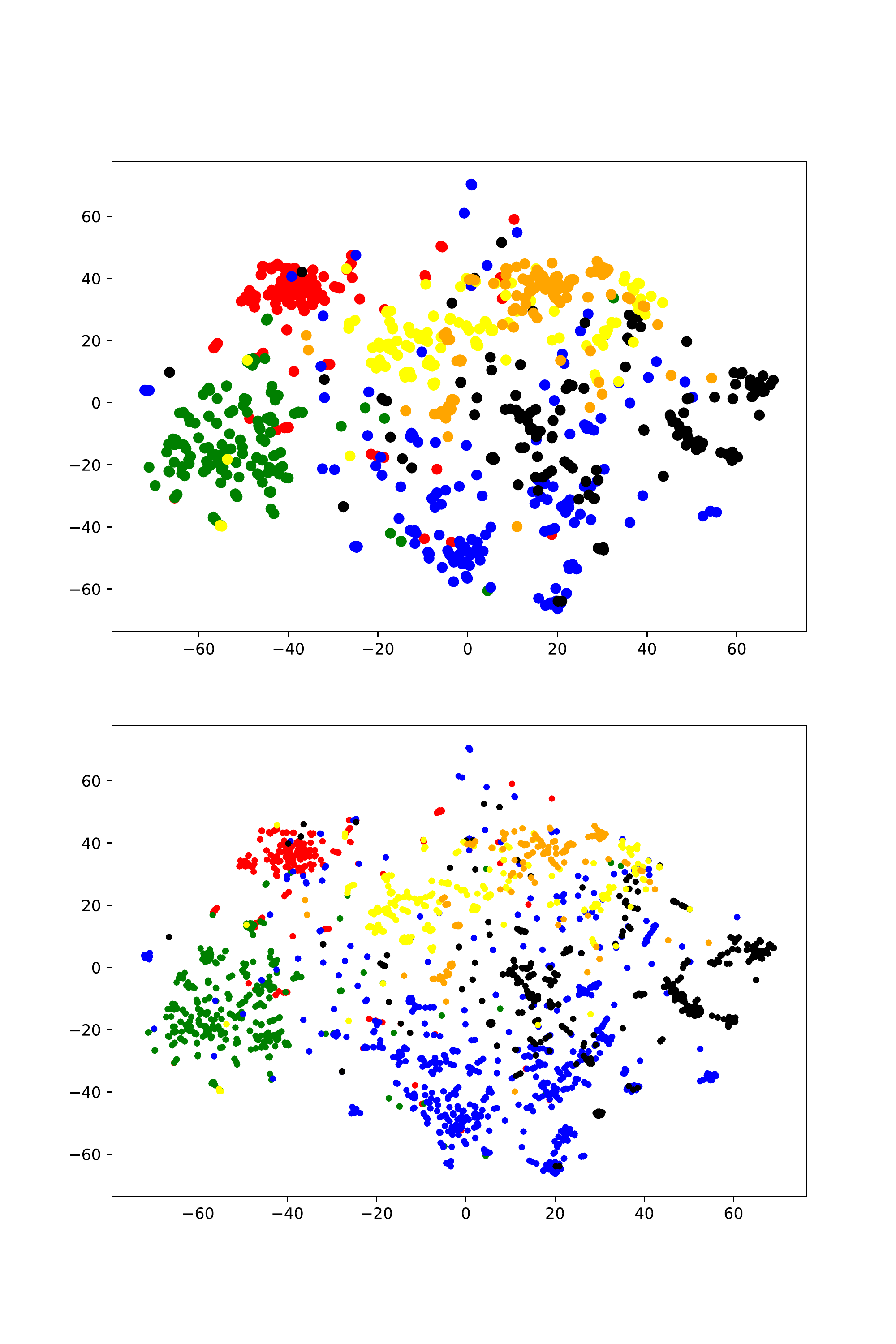}\label{fig:f1}}
\hfill
\subfloat[GraRep]{\includegraphics[width=0.15\textwidth]{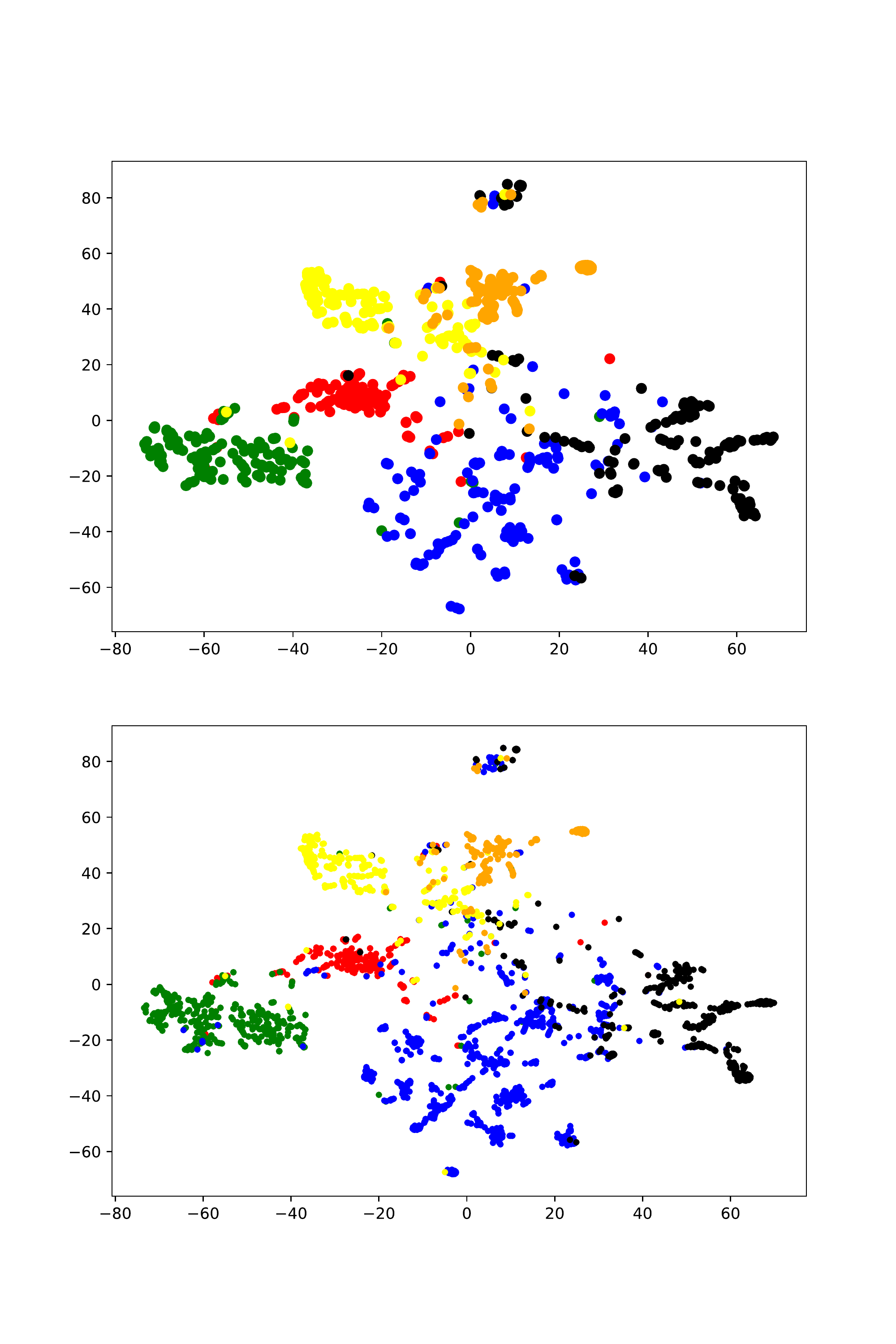}\label{fig:f2}}
\hfill
\subfloat[node2vec]{\includegraphics[width=0.15\textwidth]{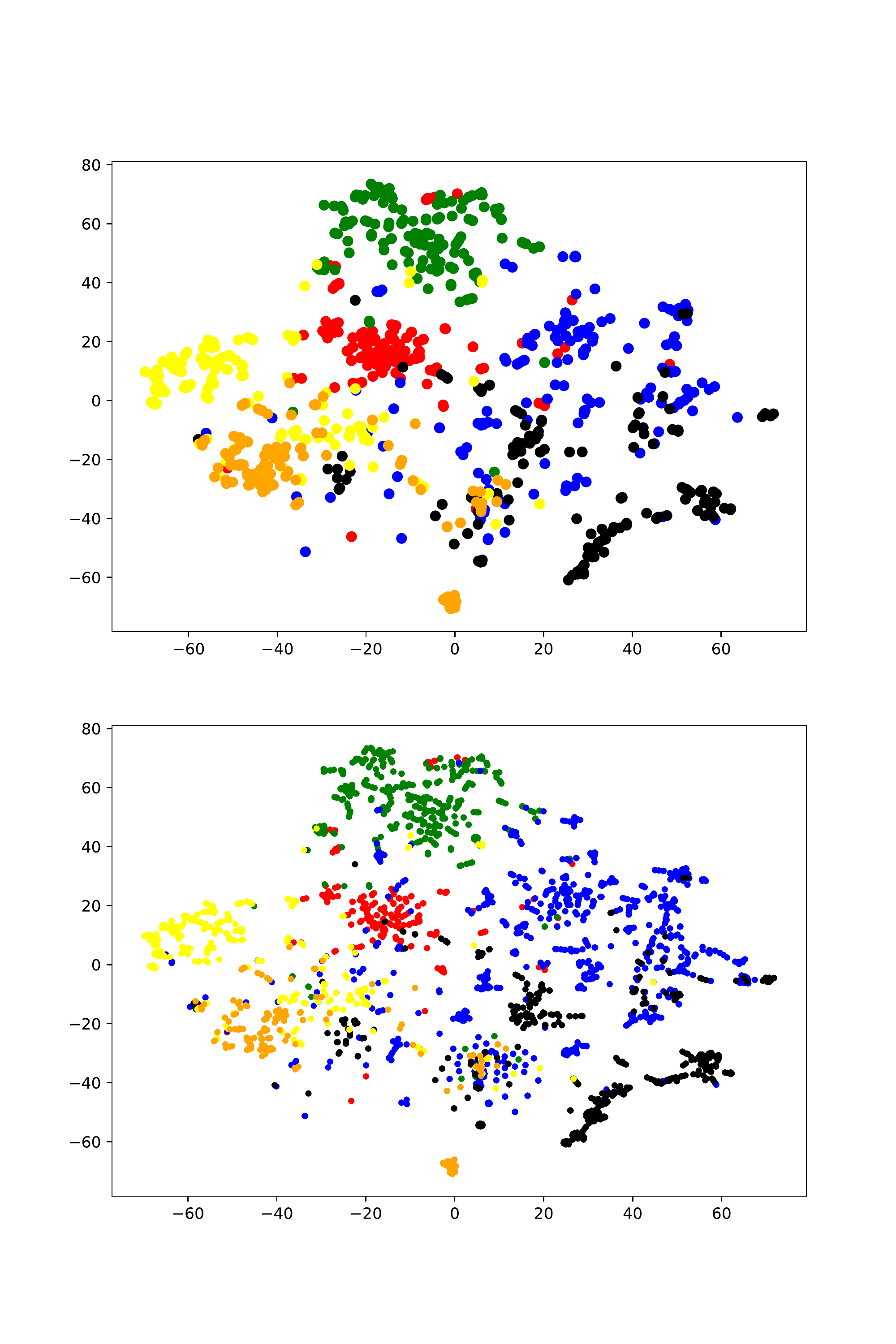}\label{fig:f2}}
\hfill
\subfloat[AIDW]{\includegraphics[width=0.15\textwidth]{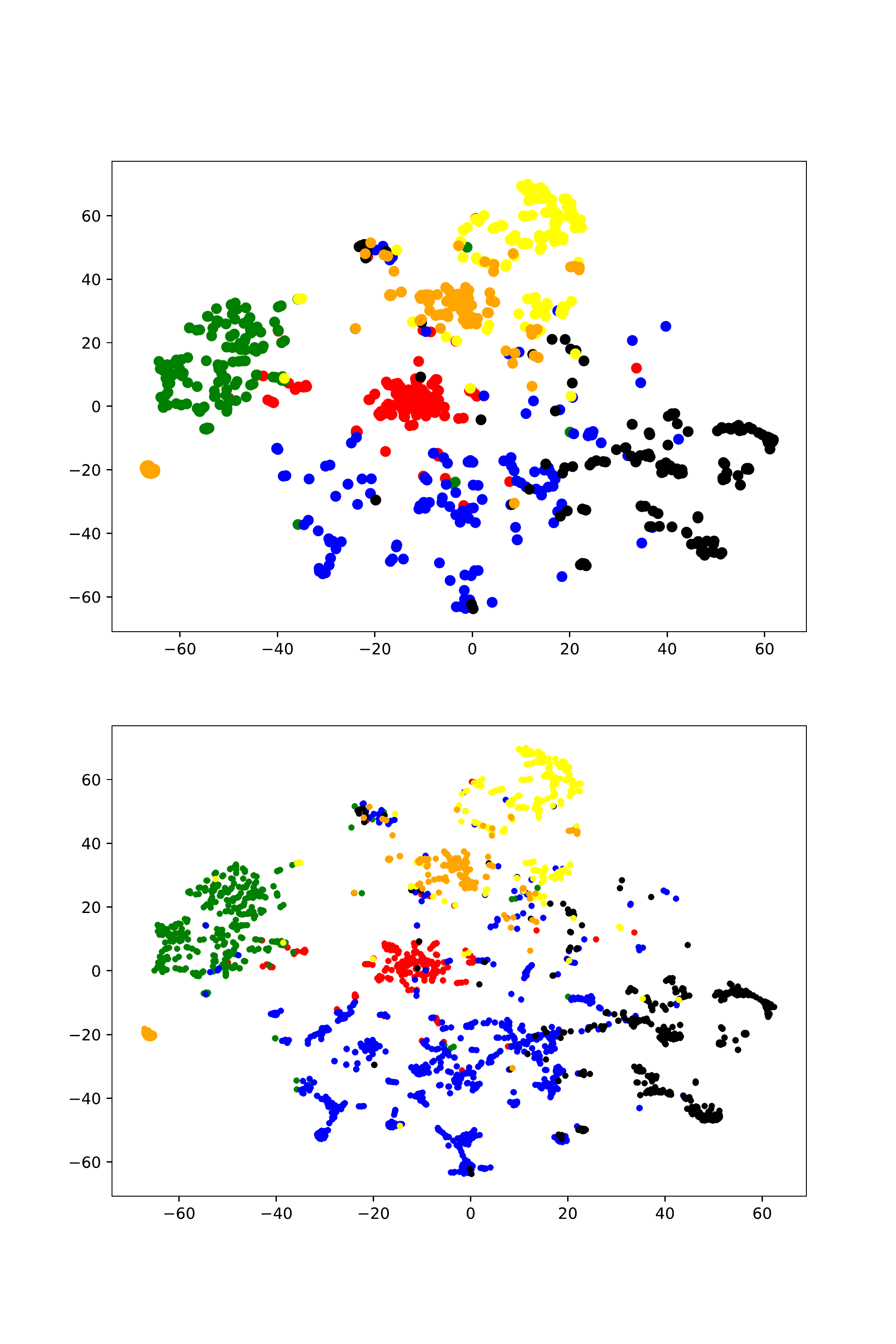}\label{fig:f2}}
\hfill
\subfloat[AAANE]{\includegraphics[width=0.15\textwidth]{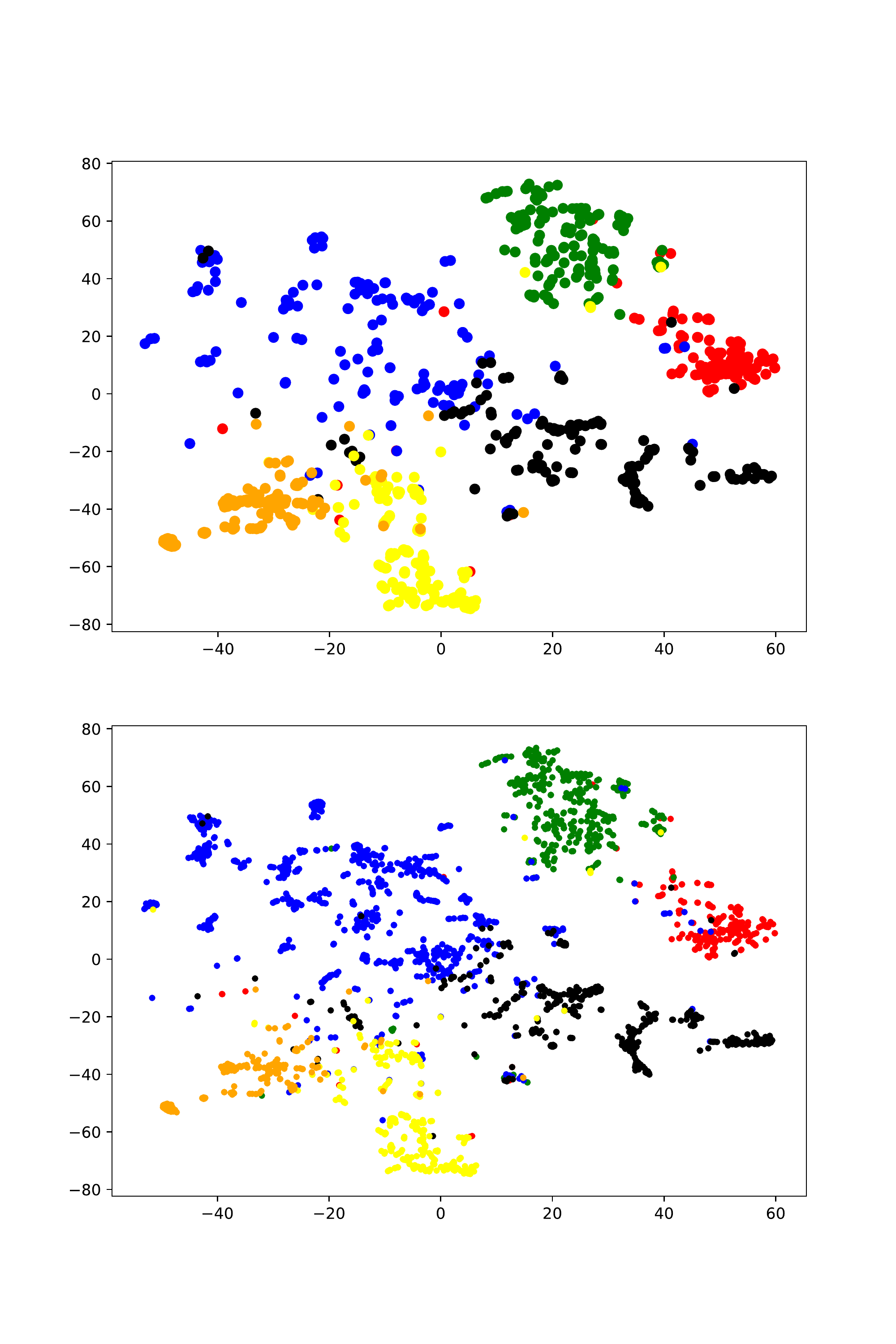}\label{fig:f2}}
\caption{Visualization of network representations learned by different algorithms on cora dataset.}
\vspace{-1mm}
\end{figure*}

\begin{figure*}[!tbp]
  \centering
  \subfloat[wiki.]{\includegraphics[width=0.47\textwidth]{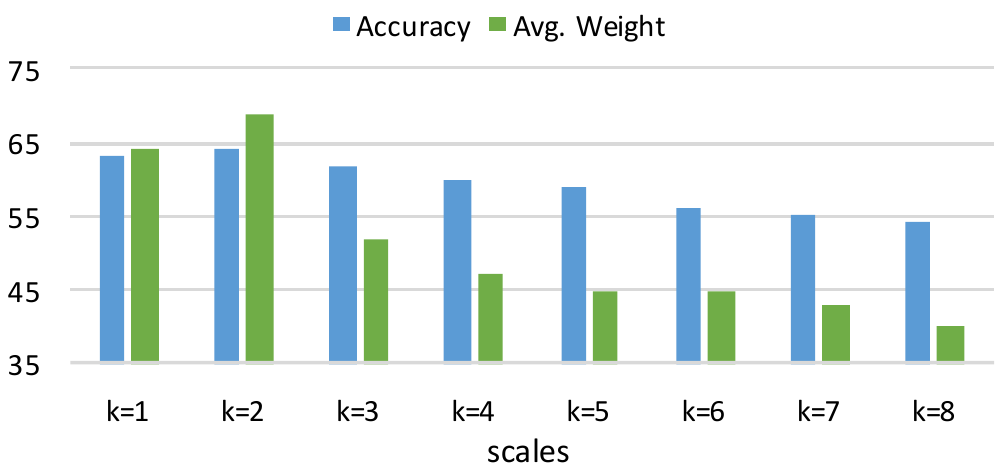}\label{fig:f1}}
  \hfill
  \subfloat[cora.]{\includegraphics[width=0.47\textwidth]{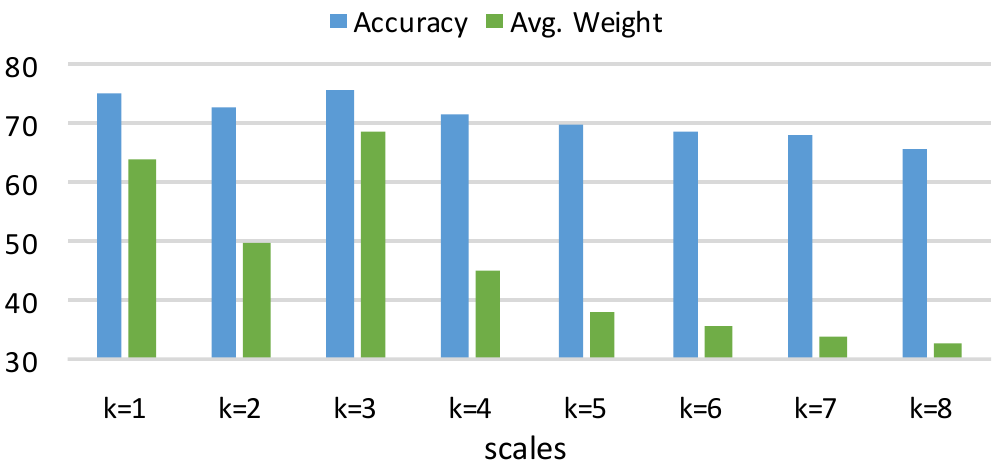}\label{fig:f2}}
\vspace*{-1mm}
  \caption{Comparison of performances on each individual scale and the average weights of scales. Scales with better performances usually attract more attentions from nodes.}
\vspace*{-2mm}
\end{figure*}

\vspace{-5mm}
\begin{figure}[]
\centering
 \subfloat[\#Dimension.]{\includegraphics[width=0.24\textwidth]{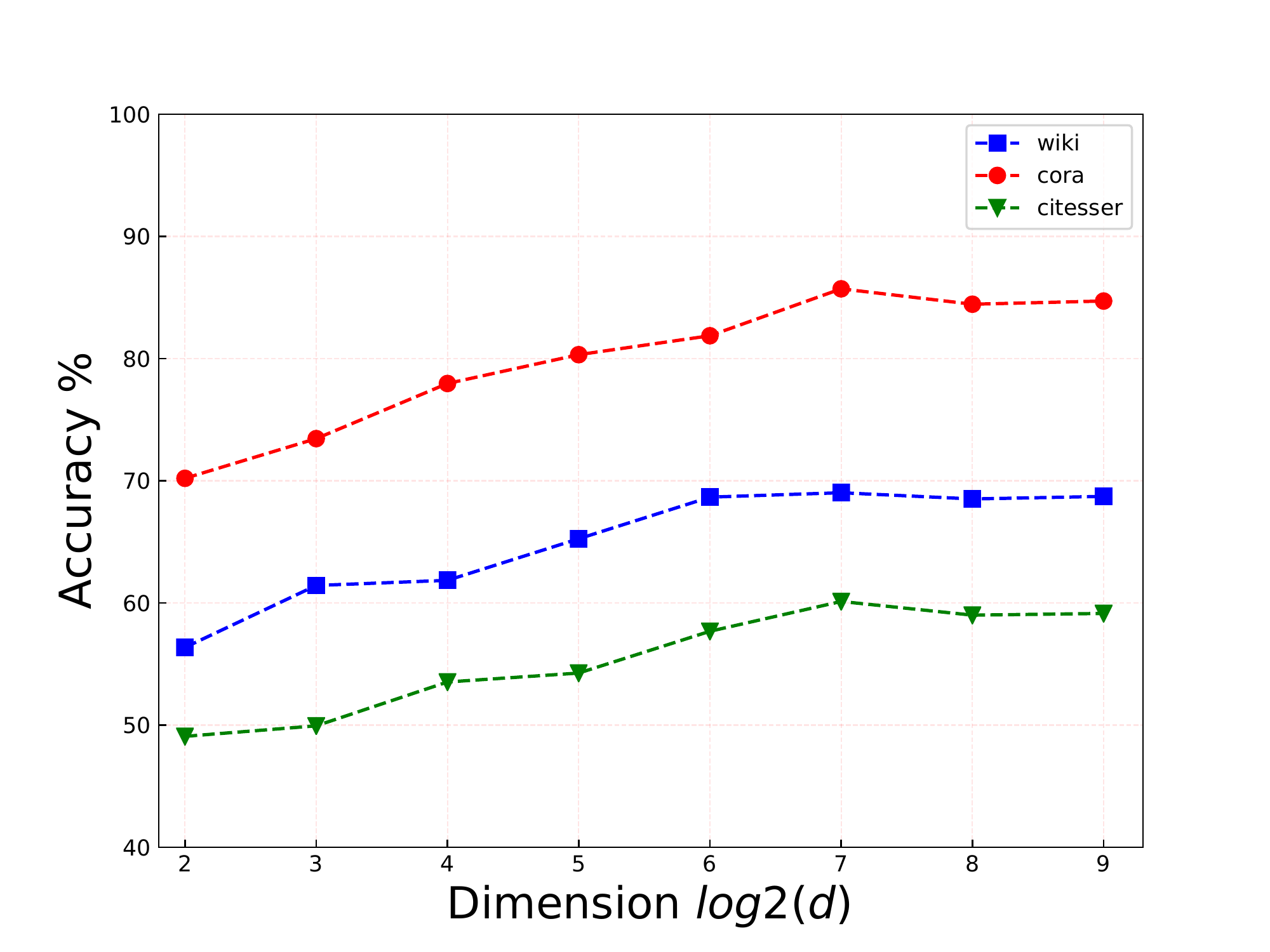}\label{fig:f1}}
 \hfill
 \subfloat[\#Scales.]{\includegraphics[width=0.24\textwidth]{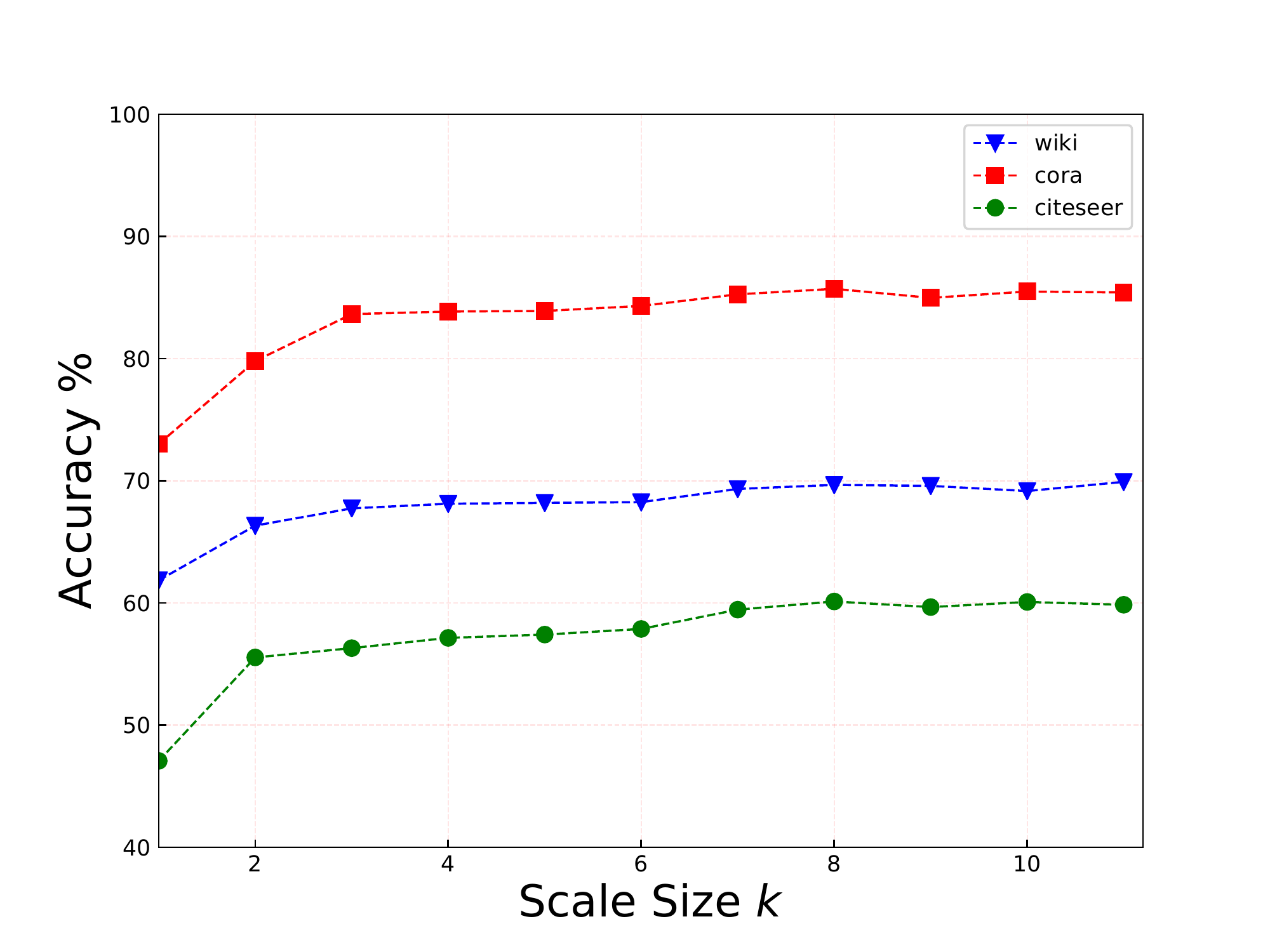}\label{fig:f2}}
 \vspace{-1mm}
\caption{Parameter sensitivity of  dimension $d$ and scale size $k$.}
\end{figure}

\vspace*{2mm}
\subsection{Baselines and Experimental Settings}
We consider a number of baselines to demonstrate the effectiveness and robustness of the proposed AAANE algorithm. For all methods and datasets, we set the embedding dimension $d = 128$.

\textbf{DeepWalk}  \cite{perozzi2014deepwalk}:  DeepWalk first transforms the network into node sequences by truncated random walk, and then uses it as input to the Skip-gram model to learn representations.


\textbf{LINE}  \cite{tang2015line}: LINE can preserve both first-order and second-order proximities for the undirected network through modeling node co-occurrence probability and node conditional probability.

\textbf{GraRep}  \cite{cao2015grarep}: GraRep preserves node proximities by constructing different $k$-order  transition matrices.

\textbf{node2vec} \cite{grover2016node2vec}: node2vec develops a biased random walk procedure to explore the neighborhood of a node, which can strike a balance between local properties and global properties of a network.

\textbf{AIDW}  \cite{Dai2017dAdversarial}: Adversarial Inductive DeepWalk (AIDW)  is an Adversarial Network Embedding (ANE) framework, which leverages random walk to sample node sequences as the structure-preserving component.



In our experimental settings, we vary the percentage of labeled nodes from $10\%$ to $90\%$ by an increment of $10\%$ for each dataset. For  DeepWalk, LINE, GraRep, node2vec, we directly use the implementations provided by  OpenNE toolkit\footnote{https://github.com/thunlp/OpenNE}. For our methods AAANE, the maximum matrix transition scale is set to 8, and the number of negative samples per input sample $m$ is set to 7. For attention-based autoencoder, it has three hidden layers, with the layer structure as $512-128-512$. For the discriminator of AAANE,  it is a three-layer neural network, with the layer structure as 512-512-1.  And the prior distributions are Gaussian Distribution following the original paper  \cite{makhzani2015adversarial}.


\vspace*{-2mm}
\subsection{Multi-label Classification}
Table 1, Table 2 and Table 3 show classification accuracies with different training ratios on different datasets, where the best results are \textbf{bold-faced}. In these tables, AANE denotes our model AAANE without Adversarial component.  From these tables, we have the following observations:
\vspace*{-1mm}
\begin{enumerate}[(1)]
\item  The proposed framework,  without leveraging the adversarial regularization version AANE, achieving average $2\%$ gains over AIDW on cora  and wiki when varying the training ratio from $10\%$ to $90\%$ in most cases,  and slightly better result on Citeseer, which suggests that assigning different weights to different scales of a node may be beneficial.
\vspace*{-2mm}
\item  After introducing Adversarial component into AANE, our Method AAANE can achieve further improvements over all baselines. It demonstrates that adversarial learning regularization can  improve the robustness and discrimination of the learned representations.
\vspace*{-2mm}
\item AAANE consistently outperforms all the other baselines on all three datasets  with different training ratios. It demonstrates that attention-based weight learning together with the adversarial  regularization can significantly  improve the robustness and discrimination of the learned embeddings

\end{enumerate}

\vspace*{-2mm}
\subsection{Visualization}
\vspace*{-1mm}
Another important application for network embedding is to generate visualizations of a network on a two-dimensional space. Therefore, we visualize the learned representations of the Cora network using the $t$-SNE package  \cite{maaten2008visualizing}.  For documents in cora dataset, which are labeled as different categories, we use different colors on the corresponded points. Therefore, a satisfactory visualization result is that the points of the same color are close to each other. From Figure 3, we observe that the representations learned by baseline methods tend to mix together. On the contrary, AAANE learns a better clustering and separation of the vertices,  and the boundaries of each group become much clearer. This  improvement proves the effectiveness of our discriminative learning model.


\vspace*{-2mm}
\subsection{Analysis of the Learned Attentions Over Scales}
In our proposed AAANE model, we adopt an attention based approach to learn the weights of scales during voting, so that different nodes can focus most of their attentions on the most informative scales. The quantitative results have shown that AAANE achieves better results by learning attention\footnote{The term attention and the term weight have the same meaning here.}over scales. In this part, we will examine the learned attention to understand why it can help improve the performances.

We  study which scale turn to attract more attentions from nodes. We take the Cora and Wiki datasets as examples. For each scale, we report the results of the scale-specific embedding corresponded to this scale, which achieves by taking only one scale vector $A^k$ as an input of autoencoder. Then, we compare this scale-specific embedding with the average attention values learned by AAANE. The results are presented in Figure 4. Overall, the performances of single scale and the average attention received by these scales are positively correlated. In other words, our approach can allow different nodes to focus on the scales with the best performances, which is quite reasonable.

\vspace*{-2mm}
\subsection{Parameter Sensitivity}
We discuss the parameter sensitivity in this section. Specifically, we assess  how the different choices of the maximal scale size K, dimension $d$ can affect node classification with the training ratio as $50\%$.

Figure 5(a)  shows the accuracy  of AAANE over different settings of the dimension $d$. The accuracy shows an apparent increase at first. This is intuitive as more bits can encode more useful information in the increasing bits. However, when the number of dimensions continuously increases, the performance starts to drop slowly. The reason is that too large number of dimensions may introduce noises which will deteriorate the performance. Figure 5(b) shows the accuracy scores over different choices of $K$. We can observe that the setting $K=2$ has a significant improvement over the setting $K=1$, and $K=3$ further outperforms $K=2$. This confirms that different $k$-order can learn complementary local information. When $K$ is large enough, learned $k$-order relational information becomes weak and shifts towards a steady distribution.
\vspace*{-3mm}
\section{Related Work}
 To preserve multi-scale structure information, some random walk and matrix factorization methods   \cite{perozzi2014deepwalk,cao2015grarep,perozzi2017don} have been proposed. GraRep \cite{cao2015grarep} accurately calculates $k$-order proximity matrix, and computes specific representation for each $k$ using SVD based dimension reduction method, and then concatenates these embeddings.  Another line of the related work is deep learning based methods. SDNE \cite{wang2016structural} , DNGR   \cite{cao2016deep}  utilize this ability of deep autoencoder to generate an embedding model that can capture non-linearity in graphs.   AIDW  \cite{Dai2017dAdversarial} proposes an adversarial network embedding  framework, which leverages the adversarial learning principle to regularize the representation learning. However, existing approaches usually  lack  weight learning for different scales.

Our work is also related to the attention-based models. Rather than using all available information, attention mechanism aims to focus on the most pertinent information for a task and has been applied to various tasks, including machine translation  \cite{bahdanau2014neural,luong2015effective} and sentence summarization  \cite{rush2015neural,shen2017disan} and question answering \cite{hermann2015teaching}.  MVE \cite{qu2017attention} proposes a multi-view network embedding, which  aims to infer robust node representations across different networks. 



\vspace*{-3mm}
\section{Conclusion and Future Work}

In this paper, we study learning node embedding for networks with multiple scales. We propose an effective framework to let different scales collaborate with each other and vote for the robust node representations. During voting, we propose an attention-based autoencoder to automatically learn the voting weights of scales while preserving the network structure information in a non-linear way. Besides, an Adversarial regularization is introduced to learn more stable and robust network embedding. Experiments on node classification and visualization tasks demonstrate the superior performance of our proposed method in learning effective  network embedding.


\bibliographystyle{named}
\bibliography{ijcai18}


\end{document}